\documentclass[lettersize,journal]{IEEEtran}
\usepackage{amsmath,amsfonts}
\usepackage{algorithmic}
\usepackage{algorithm}
\usepackage{array}
\usepackage[caption=false,font=normalsize,labelfont=sf,textfont=sf]{subfig}
\usepackage{textcomp}
\usepackage{stfloats}
\usepackage{url}
\usepackage{verbatim}
\usepackage{graphicx}
\usepackage{cite}
\usepackage{hyperref}
\usepackage{calligra}
\usepackage{algorithm}
\usepackage{amsmath}
 \usepackage{algorithmic}

\begin{document}

\title{A rule-general abductive learning by rough sets}

\author{Xu-chang Guo,~Hou-biao Li} 



\maketitle

\begin{abstract}
In real-world tasks, there is usually a large amount of unlabeled data and labeled data. The task of combining the two to learn is known as semi-supervised learning. Experts can use logical rules to label unlabeled data, but  this operation requires manual participation and requires high accuracy of expert knowledge, resulting in high costs. The combination of perception and reasoning has a good effect in processing such semi-supervised tasks with domain knowledge. However, acquiring domain knowledge and the correction, reduction and generation of rules remain complex problems to be solved. Rough set theory is an important method for solving knowledge processing in information systems. To reduce the dependence on experts for rule knowledge acquisition, correction, reduction, and generation, we propose a rule general abductive learning by rough set (RS-ABL). By transforming the target concept and sub-concepts of rules into information tables, rough set theory is used to solve the acquisition of domain knowledge and the correction, reduction and generation of rules at a lower cost. This framework can also generate more extensive negative rules to enhance the breadth of the knowledge base. {\textit{Code Availability}}: \url{https://github.com/nkjmnkjm/RS-ABL/tree/master}.
\end{abstract}


\section{Introduction}
\IEEEPARstart{W}{ith} the development of the artificial intelligence(AI), its applications in daily life are becoming increasingly widespread. However, people’s tolerance for its inexplicability is decreasing, and the demand for its generalization ability is increasing. For example, deep learning can successfully complete image classification tasks, but it requires a large amounts of labeled data and produces a module with low interpretability. The logical reasoning gets a better interpretability, but in real world, getting enough labeled data is also a time-consuming and laborious work. Therefore, bridging deep learning and logical reasoning is significant for the development of AI to solve this problem [1].

Abductive learning (ABL) [2] is a usefull framework that consists of a first-order logical reasoning model and a deep learning model. The deep learning model is used to convert the input data into symbolic data as the data for logical reasoning model; the first-order logical reasoning model reasons out the symbolic data based on the knowledge base and conducts the deep learning model. Due to the dependence of logical reasoning model on knowledge base, a better knowledge base can lead to better inference results, thus better guiding machine learning models. However, due to the difficulty of obtaining comprehensive knowledge base, how to correct, reduce and enrich knowledge base and new rules generated by logical models is an important issue.

Generally speaking, we define a rule as a combination of several sub-concepts and target concepts [3], where the target concept is the result of the rule, and the sub-concept is one of the reasons for the deducible result. For example, the concept of grandfather consists of several sub-concepts, such as the concepts of father and son.  After defining each rule as a combination of multiple concepts, we can determine whether each concept meets the definition as a label.Then, to learn target concepts, we could determine whether the sub-concept labels match the required combination of the target concept.

Rough set is an important method for determining label combinations. Since the basic concept of rough sets was proposed by Polish mathematician Z.Pawlak in 1982 [4], rough sets have become a mathematical tool for characterizing incompleteness and uncertainty. It can effectively analyze various incomplete information such as imprecision, inconsistency, and incompleteness [5,6,7]. IT can also analyze and reason data, discover hidden knowledge, and reveal potential laws [8,9,10].

In this paper, we propose a rule-general abductive learning by rough sets (RS-ABL) to correct, reduce and enrich knowledge base and new rules generated by logical models. Given a learning task, RS-ABL firstly convert rules related to the target concept into a rule table with multiple sub-labels. Then based on the rule table, RS-ABL corrects, reduces and enriches knowledge base. Afterwards, like ABL [2], the machine learning module and logic reasoning module are learned, and the new rules are processed through the rule processor based on rough set theory. After multiple empirical studies on object recognition tasks, it has been shown that RS-ABL can discover more concise sub-concept combinations to explain recognition results, outperform deep learning models in situations with less data, and generate more generalized negative concepts to guide subsequent learning.

We summarize our main contributions as follows:
\begin{itemize}
\item{We introduce rule-general abductive learning by rough set to correct, reduce and enrich knowledge base and new rules generated by logical models.}
\item{We introduced negative rules to generate logical rules with stronger generalization ability.}
\item{We evaluated the accuracy and generalization of ABL through semi-supervised experiments.}
\end{itemize}

\section{Related Work}
Integrating machine learning with logical reasoning is an important field of artificial intelligence (AI). The machine learning part is responsible for perceiving from the natural environment, while the logical reasoning part induces from symbol labels [2].

However, due to the simultaneous introduction of both learning and reasoning, most studies prioritize one part over the other [27,28]. For example, probabilistic logic program (PLP) [11] attempts to guide learning based on first-order logic, placing more emphasis on the effectiveness of reasoning and cannot fully utilize the effectiveness of learning; by contrast, statistical relational learning (SRL) [12] places more emphasis on the effectiveness of learning and cannot fully utilize the effectiveness of reasoning.

A new framework named Abductive Learning by Zhou and Dai aims to bridge machine learning and logical reasoning [2]. ABL can perform machine learning and logical reasoning simultaneously, while achieving mutual supervision between the two. Since its proposal, many researchers have conducted more in-depth and extended research on it, such as the Grounded Abductive Learning (GABL) [13]. GABL relies solely on the basic knowledge base rather than any first-order inference rules to achieve abduction. Semi-Supervised Abductive learning (SS-ABL) [14] combines abductive learning with semi-supervised learning. SS-ABL uses a small portion of labeled data and a large amount of unlabeled data to achieve the theft judical setencing. However, the existence of steps to correct specific facts in ABL raises the issue of the quality of correction. Based on this issue, Abductive Learning with Similarity (ABLsim) [15], based on the similarity of similar samples, selects better correction facts through similarity between samples.

In recent years, there has been increasing attention on how to make ABL more proactive in learning rules. Lin and Qu proposed a framewoke named Smarter Abductive Learning (S-ABL) [16], which not only summarizes rules, but also simplifies, filters, and enhances additional rules. Processing rules can better improve the quality of correcting facts. Abductive Subconcept Learning (ASL) [17] by dividing target concepts into combinations of sub-concepts, employs meta-interpretive learning (MIL) [18] to derive the relationship between sub-concepts and target concepts.

Summarize previous work, better rules can lead to better ABL results. Rough set theory is also an important method for learning, simplifying, and correcting rules. Rough set theory, as a mathematical tool and method for describing the expression, learning, and induction of uncertain and incomplete knowledge and data, can effectively analyze and process various types of information such as imprecision, incompleteness, and inconsistency, and uncover hidden knowledge and reveal potential laws from it.

Attribute reduction is one of the important steps in the knowledge discovery process, which is to maintain the classification ability of the knowledge base under unchanged conditions, delete irrelevant or unimportant attributes. Because the indiscernible binary relation is too strict in reality, scholars have also proposed many new concepts of rough sets in order to solve practical problems, such as Kryszkiewicz proposed a rough set approach to reasoning in incomplete information systems [8,9], Jenshen and Shen proposed a feature selection mechanism based on ant colony optimization to solve the problem of finding the optimal feature subset in fuzzy rough data reduction[33], Chaudhuri et al. proposed a hybrid multivariate selector method[34], Hu et al. combined Gaussian kernel with fuzzy rough set and proposed a fuzzy rough set model based on Gaussian kernel approximation[35], Greco proposed rough approximation by dominance relations [19,20].

In order to improve the computational time-consuming problem of fuzzy rough set methods, Qian et al. proposed a method for fuzzy rough feature selection accelerator[36]. And in order to deal with the problem of noise in real data, Ziarko's Variable Precision Rough set[21] provided a classification strategy with an error rate lower than the predetermined value, making the algorithm have a certain degree of fault tolerance.

\section{Preliminary}
This section is used to introduce the basic knowledge about abductive learning and rough sets.
\subsection{Abductive Learning}
The basic framework of ABL includes three parts - machine learning, logical abduction model, and consistent optimization.
\begin{itemize}
\item{Machine Learning: Learn from a sequence of samples, generate pseudo labels that can be used for abduction, and learning based on the results of abduction and consistent optimization. For the initial classifier, the accuracy is not required, and simple unsupervised learning can also be carried out.}
\item{Logical Abduction Model: Abduction (abductive reasoning) [22,23] is the process of inferring in a hypothetical space to obtain the logical explanation that best fits the result. It is carried out based on the pseudo label input in the machine learning part and the knowledge base expressed in the first-order logic language to determine whether the pseudo label needs to be corrected. The results are sent to the optimization part for pseudo label correction, and can make a judgment on the authenticity of observed facts.}
\item{Consistent Optimization: To maximize the consistency between the pseudo labels of the samples and the knowledge base, ABL needs to correct the pseudo labels to achieve consistency in the abductive results when the machine learning model does not converge. Due to the non convexity of the optimization objective, the optimization part uses derivative-free optimization to solve the problem. The corrected pseudo labels are then introduced into the machine learning part to continue learning.}
\end{itemize}

\subsection{Rough set theory}
Rough set refers to dividing a set into several equivalent classes, where each equivalent class is a rough set.

Let $U\neq\emptyset$  be a finite set composed of all the research objects, called the universe. Any subset $X\subseteq U$ is called a concept or category in $U$. Let $\textbf{R}$ be a set of equivalence relation on $U$, and for any $R\subseteq \textbf{R}$, $AS = \langle U, R \rangle$ is called an approximation space.

Let $U$ be a universe, $R\subseteq \textbf{R}$, for any $X\subseteq U$, the lower approximation ($\underline{R}(X)$) and upper approximation ($\overline{R}(X)$) of $X$ based on equivalence relation $R$ are defined as follows [24]:
\begin{equation}
    \underline{R}(X) = \{x\subseteq U:[x]_{R}\subseteq X\},
\end{equation}
\begin{equation}
    \overline{R}(X) = \{x\subseteq U:[x]_{R}\cap X \neq\emptyset\}.
\end{equation}
$[x]_{R}=\{y\subseteq U:(x,y)\subseteq R\}$ represents the equivalence class of element $X$ defined by equivalence relation $R$ on universe $U$.

Based on the definition of upper and lower approximations, we have obtained the positive region, negative region and boundary region of $X$ as follows:
\begin{equation}
    POS_{R}(X)=\underline{R}(X),
\end{equation}
\begin{equation}
    NEG_{R}(X)=U - \overline{R}(X),
\end{equation}
\begin{equation}
    BND_{R}(X)=\overline{R}(X)-\underline{R}(X).
\end{equation}

Rough set is a commonly used method for handling rule reduction and can also be used to handle rule correction and generation tasks. When we divide the target concept into combinations of sub-concepts, the data constitutes a decision information system $IS = \langle U, AT, V, f \rangle$, where $U$ is the set of elements in the data called the universe, $AT$ is the set of sub-concepts and target concepts, $\forall a\subseteq AT$, $V_a$ is the value domains of $a$, $V$ is the set of value domains of all concepts, and $f:U\times A\rightarrow V$ is an information function, $\forall x\subseteq U, a\subseteq AT$, define $f(x,a)$ to represent the value of $x$ on $a$, then $f(x,a)\subseteq V_a$.

\section{Rule-general Abductive Learning \\by Rough Set }
In this section, we introduced RS-ABL framewoke. We have provided a detailed introduction to its basic framework, rule processing by rough set and negative rule learning.
\begin{figure*}[!t]
\centering
\includegraphics[width=6.0in]{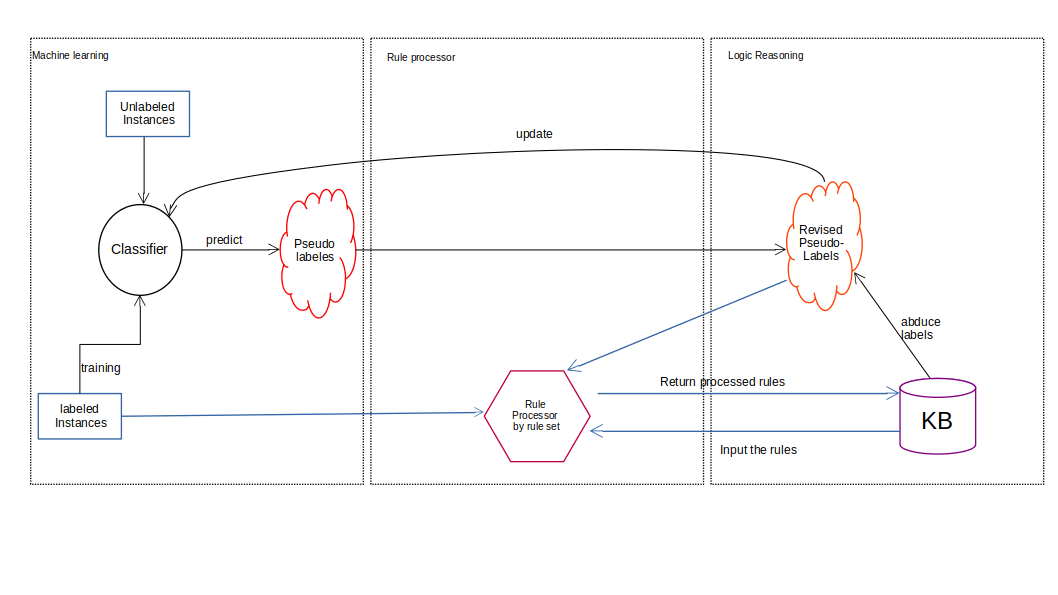}
\caption{This figure shows the basic framework of RS-ABL. The black arrow in the figure represents the basic learning process in ABL, while the blue arrow represents the newly added rule processing process in RS-ABL.}
\label{fig_1}
\end{figure*}

\subsection{Basic Framework}
As shown in Figure 1, RS-ABL consists of three parts: machine learning, rule processor, and logical reasoning. Compared to ABL, RS-ABL adds a rule processor module for correcting, reducing and enriching the knowledge base based on the labeled instances and revised pseudo-labels.

As shown in Algorithm 1, when using RS-ABL to process semi-supervised learning tasks, the input data includes a small amount of labeled data ($X_{l},Y_{l}$), a large amount of unlabeled data ($X_{u}$), and a knowledge base ($KB$). Firstly, RS-ABL trains the initial classifier with a small amount of labeled data, and corrects, reduces, and enriches the rules in the knowledge base based on this portion of labeled data. Afterwards, the unlabeled data is input into the initial classifier, resulting in a large number of pseudo labels. At this point, the pseudo label may be incorrect and needs to be processed through logical reasoning to obtain a correction that is more consistent with the knowledge base. After obtaining the revised pseudo-label, it is used for updating the classifier. At the same time, it is input into the rule processor and continue to correct, reduce and enrich the knowledge base based on the labeled data and revised pseudo-label. After updating the classifier and knowledge base, repeat the above process. Finally, we obtained a classifier and an updated knowledge base.

Compared with the original classifier, the new classifier better utilizes a large amount of unlabeled data and has better classification performance. The updated knowledge base has added corrected original rules, Simplified original rules, and generated new rules compared to the original knowledge base. And the rules in it have characteristics such as better generalization, conciseness, and better data consistency.

The goal of RS-ABL is consistent with abductive learning[2], which is to obtain a hypothetical model H that includes a recognition model P and knowledge models KB. The optimization problem of RS-ABL can be divided into two aspects: one is the optimization problem of classifiers, and the other is the optimization problem of knowledge bases. Due to the need to evaluate the accuracy of both models simultaneously, the optimization objective can be expressed as the high consistency between the results of the recognition model and the knowledge base and labeled data, while minimizing the number of corrected pseudo labels. Therefore, the following optimization function is obtained:
\begin{equation}
\begin{gathered}
\operatorname{argmin}\left(con\left(Y_l, f\right)+con\left(\Delta\left(Y_u^{\prime}\right), f^{\prime}\right)-notcon\left(Y_u^{\prime}, KB\right)\right)
\end{gathered}
\end{equation}
where $f$ means the classifier's output of labeled data, $f'$ means the classifier's output of unlabeled data, $Y'_{u}$ is the pseudo label of the unlabeled data, $\Delta(Y'_{u})$ is the revised pseudo-label by logic reasoning. $con$ is a consistency function that reflects the consistent number of two elements.

Due to the fact that the model is divided into a classifier model and a knowledge base model, the optimization process adopts an alternating optimization method, which involves alternately executing a fixed strategy and optimizing the other.

\begin{algorithm}[htb]
\renewcommand{\algorithmicrequire}{\textbf{Input:}}
\renewcommand{\algorithmicensure}{\textbf{Output:}}
\caption{ Rule-general Abductive Learning by Rough Set}
\label{alg:Framwork}
\begin{algorithmic}[1] 
\REQUIRE ~~\\ 
    Labeled data $X_{l},Y_{l}$, Unlabeled data $X_{u}$, parameter $\theta$,\\ Knowledge base $KB$, Epoch $E$
\ENSURE ~~\\ 
    Classifier $f$, New knowledge base $NKB$
    \STATE $i=0$
    \STATE $f_0 \gets TrainModel(X_{l},Y_{l})$
    \STATE $NKB_0 \gets RuleProcessor(X_{l},Y_{l},KB)$
    \STATE $While \hspace{0.1cm}i<E:$
    \STATE $\hspace{0.5cm} Y'_{u} \gets f_{i-1}(X_u)$
    \STATE $\hspace{0.5cm} \Delta(Y'_{u}) \gets Abuce(Y'_u, NKB_{i-1})$
    \STATE $\hspace{0.5cm} f_i \gets TrainModel(X_u, \Delta(Y'_{u}))$
    \STATE $\hspace{0.5cm} NKB_i \gets RuleProcessor(X_u, \Delta(Y'_{u}), NKB_{i-1})$
    \STATE $\hspace{0.5cm} \vartheta = Function(X_l, X_u, f_i, NKB_i)$
    \STATE $\hspace{0.5cm} if \hspace{0.1cm}\vartheta > \theta:$
    \STATE $\hspace{1.0cm} break$
    \STATE $\hspace{0.5cm} i = i+1$
\STATE \textbf{return} $f_i, NKB_i$
\end{algorithmic}
\end{algorithm}

\subsection{Rule processing by rough sets}
Knowledge reduction refers to obtaining a brief representation of knowledge by eliminating redundant knowledge without affecting the ability to express knowledge [4]. In rough set theory, it is the removal of irrelevant sub-concepts from existing rules to generate new rules, and based on the new rules, the target concept can still be successfully derived. As shown in  Algorithm 2, Assuming the decision information system $DIS=\langle U, C\cup D, V,f\rangle$, the reduction of subconcept set $C$ relative to the target concept set $D$ is a non empty subset $P$ of $C$, which satisfies:
\begin{enumerate}
    \item $\forall a \subseteq P$, $a$ is irreducible relative to $D$
    \item $POS_{P}(D)=POS_{C}(D)$.
\end{enumerate}
Then $P$ is called a reduction of the subconcept set $C$.
\begin{figure}[htb]
\centering
\includegraphics[width=3.5in]{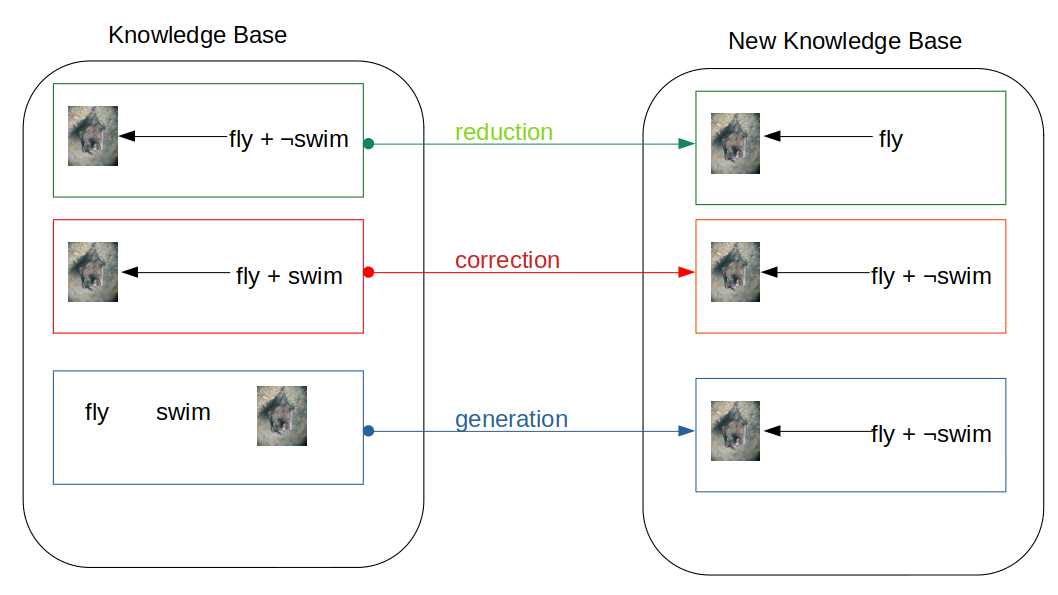}
\caption{This figure overviews three rule processing mechanisms-correction, reduction and generation.}
\label{fig_1}
\end{figure}
Rule correction is the assessment of the confidence level of a rule. When the confidence level of a rule falls below the set threshold, the rule is corrected or deleted. In the decision system $DIS=\langle U, C\cup D, V,f\rangle$, for any $x\subseteq U$, the degree of certainty ($cer(r_x)$) of the decision rule $r_{x}:des([x]_{C})\rightarrow des([x]_{D})$ corresponding to object $x$ is [4]:
\begin{equation}
    cer(r_{x})=\frac{|[x]_{C}\cap [x]_{D}|}{|[x]_{C}|}.
\end{equation}
When $cer(r_x)=1$, the rule is determinable, and when $0<cer(r_x)<1$, the rule is uncertain, then $cer(r_x)$ is the confidence level of this rule.

There are three main methods for searching hypothesis spaces in Inductive Logic Programming (ILP): Top-Down, Down-Top and Mate Level [25]. The main advantage of rough set in rule generation is that it can first reduce the sub-concepts of the information system, reduce some irrelevant or redundant information, and reduce the computational complexity of rule search.

In order to achieve subconcept reduction in information systems $DIS$, if the set family$\{X_1,X_2,...,X_n\}$ is a partition derived from the target concept $D$, then the approximate classification quality $\gamma_{C}(D)$ of $C$ for $D$ is [4]:
\begin{equation}
    \gamma_{C}(D)=\frac{|POS_{C}(D)|}{|U|}.
\end{equation}
$\gamma_{C}(D)$ is the ratio of the objects in the target concept $U/D$ that can be accurately divided under the sub-concept set $C$ to the total number of objects, indicating the degree of dependency of the sub-concept set on the target concept set. $A\subseteq C,\forall a\subseteq(C-A)$, the importance of subconcept $a$ relative to $D$ is:
\begin{equation}
    Sig(a,A,D)=\gamma_{A\cup\{a\}}(D)-\gamma_{A}(D).
\end{equation}

By using the importance of sub-concepts $Sig(a,A,D)$, the sub-concept reduction of the sub-concept set $C'$ is calculated. In the remaining sub-concept set $C'$, conventional rule search algorithms are used to obtain new rules.

\begin{algorithm}[htb]
\renewcommand{\algorithmicrequire}{\textbf{Input:}}
\renewcommand{\algorithmicensure}{\textbf{Output:}}
\caption{Rule generation algorithm}
\label{alg:Framwork}
\begin{algorithmic}[1] 
\REQUIRE ~~\\ 
     decision system $DIS=\langle U, C\cup D, V,f\rangle$
\ENSURE ~~\\ 
    new rules $R$
    \STATE $R= [\hspace{0.1cm}]$
    \STATE $for \hspace{0.1cm}i \hspace{0.1cm} in \hspace{0.1cm}D:$
    \STATE $\hspace{0.5cm} A \gets minReduction(i)$
    \STATE $\hspace{0.5cm} if \hspace{0.1cm} POS_{A}(D)==POS_{C}(D):$
    \STATE $\hspace{1.5cm} R.append(A,i)$
    \STATE $\hspace{0.5cm} else:$
    \STATE $\hspace{1.5cm} a \gets max Sig(a,A,i) (a \subseteq C)$
    \STATE $\hspace{1.5cm} A \gets A \cup \{a\}$
    \STATE $\hspace{1.5cm} until\hspace{0.1cm} POS_{A}(D)==POS_{C}(D)$
    \STATE $\hspace{2.5cm} R.append(A,i)$
\STATE \textbf{return} $R$
\end{algorithmic}
\end{algorithm}


\subsection{Negative Rule Learning}
In general rule learning, affirmative rule learning is the mainstream learning trend. In ABL, the general composition of a knowledge base is affirmative rules, that is, if-then rules. However, if-then rule sometimes does not have good extensiveness, such as in learning image recognition tasks for human and tiger, learning affirmative rules will result in the following rules: $tiger(A)\gets quadruped(A)$. But this rule loses its effect when dealing with the classification task of tiger and bear.

Compared to the if-then rule, the if-then not and if not-then not rules have better extensive, such as $nottiger(A)\gets Uprightwalking(A)$. However, general rule search algorithms require the construction of negative sub-concepts and negative target concepts in advance, which increases the cost of building a knowledge base.

Rough sets can directly generate negative rules using the same process as the rule generation part based on the negative sub-concept support set $||\neg t||=\{x\subseteq U:f(x,a)=*\land f(x,a)\ne v_a\}$ and the negative target concepts support set $||\neg s||=\{x\subseteq U:f(x,d)\ne i,i\subseteq v_d\}$ [26].

From the above definition of a negative support set, it can be seen that a negative support set is a complement to a certain partition. Therefore, learning negative rules based on rough sets does not require the reconstruction of new sub-concepts and target concepts, and only requires supplementing the original set, which greatly reduces the workload of solving negative rules. After constructing the negative support set, we can learn negative rules based on the method of rule generation.
\begin{figure}[htb]
\centering
\includegraphics[width=3.5in]{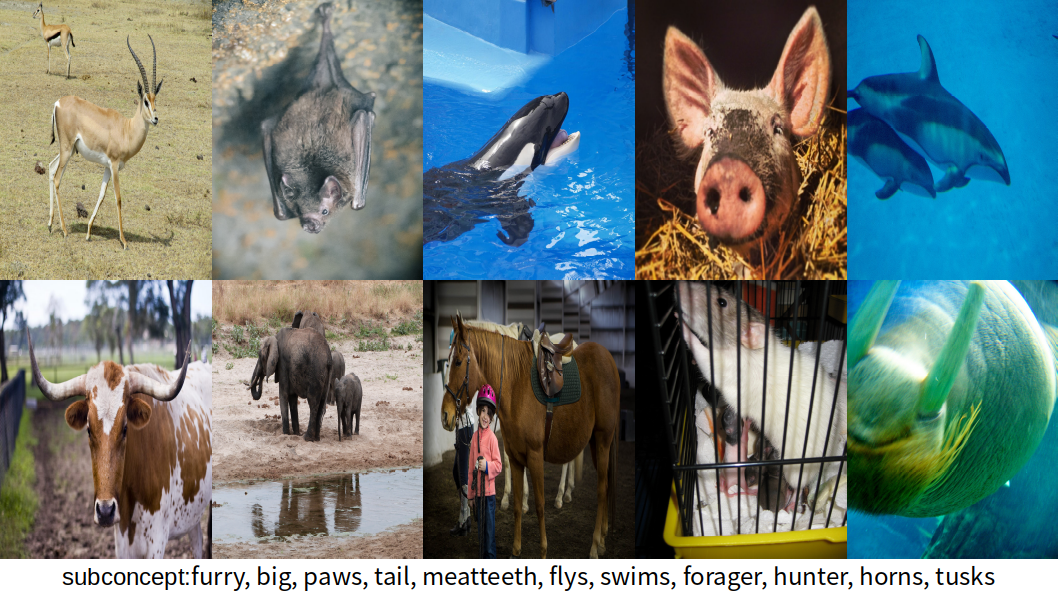}
\caption{This image shows 10 selected animal image data and 11 features.}
\label{fig_1}
\end{figure}
\section{Experiments}
In this section, the main objectives of our experiment are to 1) test whether RS-ABL can learn logical rules that conform to the dataset through information tables, 2) test whether RS-ABL can reduce, correct, and enrich the knowledge base, and 3) test the generalization ability of negative rules generated through RS-ABL.

To achieve the experimental objectives, this paper designs a semi supervised image classification task using RS-ABL and compares its performance with other ABL algorithms.
\subsection{Task}
This experiment mainly tests the ability of RS-ABL through semi-supervised tasks. The task goal is to train a classifier and a knowledge base. The classifier is used to implement the classification task of animal images; The knowledge base is used to contain new first-order logical rules generated during the training process for animal classification.

\subsection{Dataset and Experimental Setup}
The dataset in the experiment utilized the real dataset animals with attributes 2 (AWA2)[32]. The AWA2 dataset consists of 50 animals, 85 labeled features, and 37322 images. As shown in Figure 2, in this experiment, we conducted six random repeated experiments, each time selecting 10 animals, 11 features.And for each training set, randomly select 10\%, 50\%, and 100\% of the training set as labeled data, and conduct 5 independent repeated experiments in each case.
Each time, 80\% of the data is selected as the training set, and 20\% of the data is used as the test set. Five independent repeated experiments are conducted on each group of data with Intel(R) Core(TM) i7-10700F CPU@2.90HZ and GeForce RTX 2060 GPU, and the average of the five experimental results is taken.

We select common semi-supervised learning methods-MixMatch [29], FixMatch [30] as comparison items. WideResNet [31], depth=28, widen\_factor=2, is used as the same classifier of all semi-supervised learning methods. During data preprocessing, the images are uniformly sized to $32 \times 32 \times 3$, and the data is divided into labeled data and unlabeled data, labeled data accounting for 10\%, 50\%, and 100\%, respectively.Set experimental parameters batch\_size=64, epochs=15, train-iteration=30, lr =0.01,$\theta$ = 0.8.

\subsection{Knowledge Base}
The knowledge base generally predetermined consists of animal discrimination rules represented by first-order logical rules. Rules can be negative or positive. Moreover, due to the presence of a rule generator, the initial knowledge base can be empty. In each experiment of image classification task, the knowledge base contains some correct rules, incorrect rules, and repeated rules to test the ability of RS-ABL to reduce, correct, and enrich the knowledge base. For example, in an experiment, the pre-set knowledge base only added the following three rules to demonstrate the ability of RS-ABL correction, reduction, and generation of new rules, such as $1) \neg bat \gets \neg flys$; $2) \neg bat \gets \neg flys + \neg swims$; $3) bat \gets \neg flys$.
\begin{figure}[!t]
\centering
\includegraphics[width=3.5in]{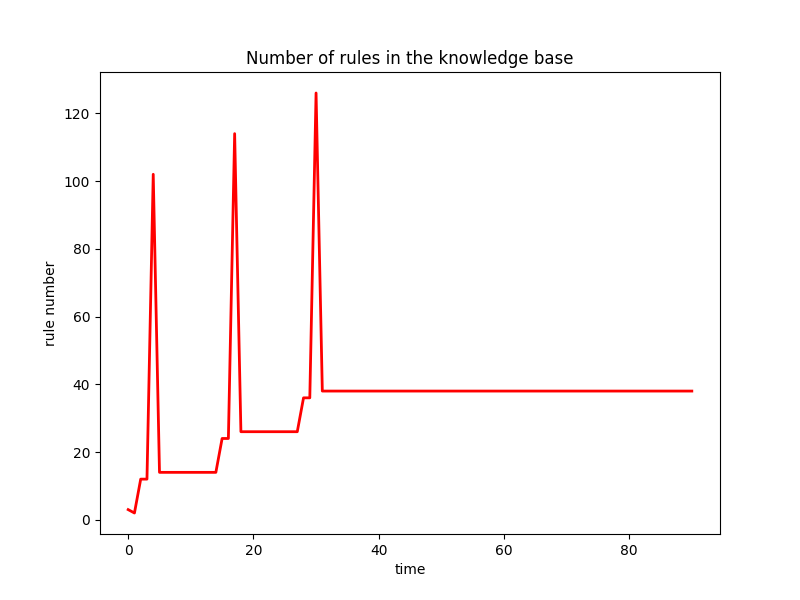}
\caption{The variation of the number of rules in the knowledge base with the running process of RS-ABL.}
\label{fig_1}
\end{figure}

\begin{table}[!t]
\caption{Top-1 and Top-5 accuracy in experiments\label{tab:table1}}
\centering
 \renewcommand{\arraystretch}{2.0}
\setlength{\tabcolsep}{7mm}
\begin{tabular}{c c c}
\hline
Methods & Top-1 acc &Top-5 acc\\
\hline
MixMatch-10 & 0.305 & 0.700 \\
FixMatch-10& 0.132& 0.718\\
RS-ABL-10 & \textbf{0.685} & \textbf{0.962}\\
\hline
MixMatch-50 & 0.34 &0.823 \\
FixMatch-50& 0.173& 0.758\\
RS-ABL-50 & \textbf{0.708} & \textbf{0.975}\\
\hline
WideResNet-100 & 0.765 & 0.987\\
RS-ABL-100 & 0.768& 0.985\\
\hline
\end{tabular}
\end{table}

\subsection{Results}
The experiment selected top 1 accuracy and top 5 accuracy to evaluate the effectiveness of the model, in order to reflect the prediction accuracy of the model. The Top1 accuracy is the probability that the model predicts the most likely results to be true, while the Top5 accuracy is the probability that the top 5 most likely results predicted by the model contain true results.

Table 1 shows the experimental results of animal classification comparison test. It can be found that the results of RS-ABL under each label rate are better than the classical semi-supervised learning method.When the label rate is low, the accuracy of RS-ABL that combines logical reasoning and machine learning is significantly higher.And it can be found that when dealing with tasks with obvious classification rules, the classification effect of RS-ABL still has a good effect when the label rate is low.

Figure 4 shows the changes in the number of rules in the knowledge base over time. It can be observed that the number of rules fluctuates, indicating the process of rule reduction, correction, and generation in the knowledge base, and ultimately achieving a balance. The number of rules in the knowledge base no longer changes.

\subsection{Theft Judicial}
The task of Table 2 is to predict the defendant's sentence.The dataset uses the judicial dataset from Semi Supervised Abductive Learning (SS-ABL) [14]. Introducing knowledge base(KB) as a comparative factor reflects the model's dependence on the KB, where FULL represents the existence of a complete KB and N/A represents the absence of a KB. We compared the Mean Absolute Error (MAE) and Mean Square Error (MSE) of SS-ABL, RS-ABL, RS-ABL with empty knowledge base and linear regression (LR) to display the predicted accuracy of the model.
From the results, it can be seen that the results of RS-ABL and SS-ABL are basically consistent when the knowledge base is strong. When the knowledge base is empty, RS-ABL can still improve the accuracy of machine learning.
\begin{table}[htb]
\caption{MAE and MSE in experiments\label{tab:table1}}
\centering
 \renewcommand{\arraystretch}{2.0}
\setlength{\tabcolsep}{5mm}
\begin{tabular}{c c c c}
\hline
KB&Methods & MAE &MSE\\
\hline
FULL& SS\_ABL-10 & 0.825 & 1.147\\
FULL& RS-ABL-10 & 0.824 & 1.138\\
N/A & LR-10 &0.894 & 1.252\\
N/A & RS-ABL-10 &0.875 & 1.239 \\
\hline
\end{tabular}
\end{table}
\section{Conclusion}
In this paper, we propose a rule-general inductive learning by rough set (RS-ABL), aimed at solving rule correction, reduction, and generation in ABL. RS-ABL utilizes rough set principles to correct, reduce, and generate rules by generating rule tables from sub-concepts and target concepts in the knowledge base, and can generate negation rules with better universality. In the contrast experiment with traditional semi-supervised learning, RS-ABL has better accuracy. And the rule requirements for the initial knowledge base are not strong, only sub-concepts and target concepts need to be constructed, which greatly reduces dependence on expert rules. In future research, it may be considered to add probability components to the rule processing process to handle incomplete information systems. And when the initial setting of the knowledge base is empty, rules learned solely from rough set theory may have adverse effects on subsequent learning. How to constrain the application of rough set theory in rule learning in labeled data is also a future research direction.\\
\indent {\bf{Data Availability}}: \url{https://cvml.ista.ac.at/AwA2/} and \url{https://github.com/AbductiveLearning/SS-ABL/tree/master/data}\\
\indent {\bf{Code Availability}}: \url{https://github.com/nkjmnkjm/RS-ABL/tree/master}

\vfill

\end{document}